\documentclass[journal]{IEEEtran}
\usepackage{graphicx}
\usepackage[english]{babel}
\usepackage[utf8]{inputenc}
\usepackage{times}
\usepackage{amssymb,amsfonts}
\usepackage[tbtags]{amsmath}
\usepackage{cite}
\usepackage{slashbox}
\usepackage{pict2e}
\usepackage{float}
\usepackage[all]{xy}
\usepackage{graphics,graphicx,color,colortbl}
\usepackage{times}
\usepackage{subfigure}
\usepackage{wrapfig}
\usepackage{multicol}
\usepackage{cite}
\usepackage{url}
\usepackage[tbtags]{amsmath}
\usepackage{amsmath,amssymb,amsfonts,amsbsy}
\usepackage{bm}
\usepackage{clrscode3e}
\usepackage[centerlast, small]{caption}
\usepackage[colorlinks=true, citecolor=blue, linkcolor=blue, urlcolor=blue, breaklinks=true]{hyperref}
\usepackage{fancyhdr} 
\makeatletter
\def\markboth#1#2{\def\leftmark{\@IEEEcompsoconly{\sffamily}\MakeUppercase{\protect#1}}%
\def\rightmark{\@IEEEcompsoconly{\sffamily}\MakeUppercase{\protect#2}}}
\makeatother

\begin{document}

\title{Named Entity Extraction with Finite State Transducers}

\author{\IEEEauthorblockN{Diego Alexander Huerfano Villalba\IEEEauthorrefmark{1},
Elizabeth León Guzmán\IEEEauthorrefmark{2}}\\
\IEEEauthorblockA{MIDAS Research Group\\ Department of Computer Engineering\\
National University of Colombia\\
Email: \IEEEauthorrefmark{1}dahuerfanov@unal.edu.co,
\IEEEauthorrefmark{2}eleonguz@unal.edu.co}
}

\maketitle

\begin{abstract}
We describe a named entity tagging system that requires minimal linguistic knowledge and can be applied to more target languages without substantial changes. The system is based on the ideas of the Brill's tagger which makes it really simple. Using supervised machine learning, we construct a series of automatons (or transducers) in order to tag a given text. The final model is composed entirely of automatons and it requires a lineal time for tagging. It was tested with the Spanish data set provided in the CoNLL-$2002$ attaining an overall $F_{\beta = 1}$ measure of $60\%.$ Also, we present an algorithm for the construction of the final transducer used to encode all the learned contextual rules.
\end{abstract}


\section{Introduction}

\IEEEPARstart{T}{he} task of Named Entity Extraction (NEE) consists in detection of named entities that appears in a given text.
Named entities are understood to be anything that has a proper name (people, places, organizations and so on) \cite{libroPLN}.
The output obtained form this process is widely used in several Natural Language Processing (NLP) applications like Question Answering, Machine Translation, Information Retrieval, Summarization and Topics Detection. This task is carried out through two main sub-tasks: Named Entity Recognition (NER) and Classification (NEC) which are often merged in a single process. This work presents a system based on finite state transducers to perform NEE where these two sub-tasks are performed simultaneously.

Approaches to NEE with finite state automatons have been previously developed. In \cite{paperSerbio} is proposed a system for NER in Serbian. Their system is based on handmade finite state automatons and E-dictionaries recognizing several kinds of names and temporal and numeric expressions. The overall performance of this system is a recall of $80\%$ and a precision of $98\%.$ On the other hand, in \cite{paperAdquisicionAutomatas}, it is presented an algorithm using the approach of Casual-States Spliting Reconstruction (CSSR) where an automaton is learned from sequential data. In the end, this automaton is able to recognize named entities in a text. The system had a recall of $88\%$ and a precision of $89\%.$

This paper presents a system for the complete NEE task using finite state transducers learned from annotated texts. The system is based on the ideas of the 
Brill's tagger \cite{paperBrill} in order to learn change rules, then a deterministic transducer is constructed from these rules. It is also presented the results and evaluation using the data from the CoNLL-$2002$ Shared Task.

\section{Description of the System} 
The system presented here uses the BIO scheme also used in the CoNLL-$2002$ in order to identify  four kinds of named entities: person, location, organization and a last class called miscellaneous. For tagging a given text we first use a lexical tagger which only takes into account the form a single word and its most likely tag according to the training text. This first stage can be seen as a initialization. Then we infer change rules to correct those initial tags considering the context which a single word appears in (words before and after that word). This is called contextual tagger. This approach was proposed in \cite{paperBrill} for the part-of-speech task. In addition, we encode the contextual tagger into a single deterministic finite state transducer allowing to tag the text with a linear complexity. The construction of such a transducer was proposed in \cite{paperTransducers}, however, we propose here a new algorithm for the \emph{local extension} procedure since we have found that the algorithm introduced there does not work in all cases. 

The training text is divided in two parts. The first one is used for the lexical tagger so that we can estimate the most likely tag of every word. The second part is used to infer change rules allowing us to test which is the best rule.

\subsection{Lexical Tagger}
The purpose of this module is to give a initial tag a every single word in a text. In the same way that Brill proposed in \cite{paperBrill}, we have for every word a register of the most likely tag according to the given training text, which sometimes can be wrong. We used the structure \emph{trie}, which can be thought as a finite state automaton which accepts a finite language, to store all words and their most likely tag allowing us to search the tag of a given word in the dictionary in linear time. Since some words can be unknown, a second trie is used to store words suffix and its most likely tag, so that we can tag unknown words using only its suffix (of 4 characters length). Also, we encode every word in the training text replacing its capital characters by 'X', lower case characters by 'x', digits by 'd' and the remaining characters are not changed. In this way, we can tag unknown words considering only their form according to known words with similar form. This information is also held in a new trie.

\subsection{Contextual Tagger}
Since the prior initialization can be wrong for some words, we need to change some initial tags to the correct ones taking into account the context where each word is. This context is, in our case, the words tags around one word (we considered up to two previous tokens and up to two next tokens). As in the Brill's tagger, we construct an ordered list of these correction rules which are applied along all the text in the given order. An example of the structure of these rules is $abt_1de \rightarrow abt_2de$ where $a,b,c,d,t_1, t_2$ can take the value of any possible tags that we are working with (according to the BIO scheme) and are considered to be variables for this structure. This example means that when we find the pattern $abt_1de$ in the text, we change the tag $t_1$ with the tag $t_2$ and the remaining tags in the pattern are left unchanged. In this way, we are changing specific tags in the text which are in a specific context. 

So as to make the list of contextual rules, we infer each rule in a stage, where every structures with all possibles values for each variable are tested with the initialized text and then it is calculated its score (it is the difference between  the number of correct changes made and the number of incorrect changes made in comparison with the training test). The rule with the best score in a stage is chosen and added to the ordered list. Then we apply it to the initialized text and we continue with the next stage.

As it is observed in \cite{paperTransducers}, apply these rules may require $RCn$ elementary steps to tag an input of $n$ words with $R$ contextual rules requiring at most $C$ tokens of context. Therefore, it is proposed there a new way to encode all these rules into a single deterministic finite state transducer allowing to do the same procedure in $n$ elementary steps, which corresponds to follow a single path in the transducer according to the tags that we find when we are processing the text token by token. In the end, we will have constructed a model which is based in automatons (or transducers) where all the tagging stage is performed in linear time.

\subsection{Trigger Words and Punctuation Marks}

In addition to the tags proposed in the BIO scheme, we also used some tags to identify trigger words (prepositions) which are likely to appear before a named entity like \textit{en}, \textit{de}, \textit{por}, \textit{según}. This is the only assumption about the Spanish language. Also, there is a tag called \textit{PUNCTUATION} which covers all the punctuation marks in the text. This was made so that contextual rules with tags of the BIO scheme apply only in complete sentences. 

\section{Construction of the Deterministic Finite State Transducer}

On overall, the construction of the final transducer, given the ordered list of contextual rules, is carried out in the following three steps:

\begin{enumerate}
\item For each contextual rule it is constructed the \textit{local extension transducer}.
\item The composition operation is applied over the transducers in the same order as their corresponding rules were in the ordered list. The result of this step is a single transducer which can be non-deterministic.
\item It is applied the determinization algorithm over the final transducer.
\end{enumerate}

This method was proposed in \cite{paperTransducers} for the part-of-speech tagging. There it is proved that the kind of rules which come up in this process can be represented by deterministic finite transducers, therefore the well known determinization algorithm for transducers finishes in some point. Since the composition algorithm for transducers is also a standard algorithm in the literature, we will focus on the explanation of the  local extension algorithm.

\subsection{Local Extension of a Rule}

In this section we show how to construct the local extension transducer of a given contextual rule $$a_1 \cdots a_n t_1 a_{n+1} \cdots a_{n+m} \rightarrow a_1 \cdots a_n t_2 a_{n+1}\cdots a_{n+m}$$
where we change $t_1$ by $t_2.$ The local extension transducer is a transducer which will be able to find all the occurrences of $a_1 \cdots a_n t_1 a_{n+1} \cdots a_{n+m}$ in a given text and replace them by $a_1 \cdots a_n t_2 a_{n+1}\cdots a_{n+m}$ leaving all the other characters outside the occurrences unchanged. In our algorithm we consider all the occurrences in a text without overlaps giving priority to the occurrence which appears first.

\begin{figure}[!t]
\centering
\includegraphics[width=2.0in]{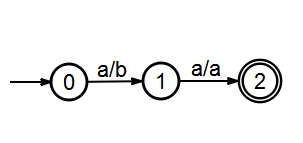}
\caption{Transducer for the rule $aa \rightarrow ba$.}
\label{fig_auto0}
\end{figure}

\begin{figure}[!t]
\centering
\includegraphics[width=2.5in]{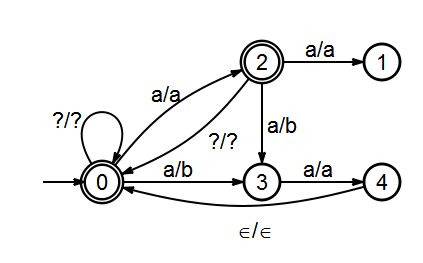}
\caption{Ambiguous local extension for the rule $aa \rightarrow ba$.}
\label{fig_auto1}
\end{figure}

As we have pointed out before, in \cite{paperTransducers} it is presented an algorithm for the same purpose with the difference that they take a given rule encoded in a transducer and then the algorithm computes the local extension from this given automaton. However, we have noticed that the given procedure leads to ambiguities. In the figure \ref{fig_auto0} it is represented a transducer encoding the rule $aa \rightarrow ba$ and in the figure \ref{fig_auto1} it is shown the result of their algorithm for this transducer. We can see that if have the text $aaa$ there are two possible outputs, which finish in final states: $aba$ with the sequence of states $0,$ $2, 3, 4, 0$ and $baa$ with the sequence of states $0, 3, 4, 0, 2.$ These kinds of ambiguities are problematic when the algorithm of determinization is applied because there is no unique output for the same sequence of characters. Besides, we have two occurrences overlapping in the example: $\mathbf{aa}a$ and $a\mathbf{aa},$ but, according to our definition of local extension, we give priority to the first found occurrence, therefore the only correct output would be $\mathbf{ba}a.$ Note that we do not apply the rule to $b\mathbf{aa}$ in the last output and obtain $b\mathbf{ba}$ since we only consider occurrences without overlaps.

The algorithm presented here to solve this problem is based on the algorithm of string matching KMP. For string matching we can construct a finite automaton that scans the text string $P$ for all occurrences. \cite{clrs} presents a method for building such an automaton which solves the problem in linear time. In our case, we need to find the pattern $P$ and replace it for a new text $P'.$ According to the form of every contextual rule, the difference between the texts $P$ and $P'$ is only a character, then the input of the algorithm is the text $P,$ an index $0\le k < P.length$ and a character $c$ which means that $P'$ is the same text $P$ but with the character $c$ at the position $k.$

We adopt the same notation used in \cite{clrs} and \cite{paperTransducers}. Given two strings $S_1, S_2$ we denote $S_1 \sqsupset S_2$ if $S_1$ is suffix of $S2$ and $S_1 \sqsubset S_2$ if $S_1$ is prefix of $S_2.$ For a given string $S$ and $-1 \le k < S.length,$ $S_k =S[0, \dots, k].$ Note that $S_{-1}$ is the empty string. Also, if $0\le k \le S.length,$ $_{k}S = S[k, \dots S.length-1].$ In the same way, $_{S.length}S$ is the empty string. Edges in a transducer are of the form $(p,a,b,q)$ which means that there is an edge from the state $p$ to the state $q$ reading the symbol $a$ and writing the symbol $b.$ A question mark ($?$) on an edge transition (for example labeled $?/b$) originating at state $p$ stands for any input symbol that does not appear as an input symbol on any other out-going edge from $p$.

The transducer can be seen in three non-disjoint parts. The first one corresponds to the processing of the prefix $P_{k-1}$ and it is constructed in the same way that the automaton for string matching is constructed and where each transduction outputs the same symbol that it reads. When we have found the prefix $P_k$ in the text, we have two options:  leave  the symbol $P[k]$ unchanged or replace it by $c$. With the first option, we would not expect to find all the pattern $P$ because in the other case we should have changed the symbol $P[k]$ by $c.$ With the second option, we would expect to find all the pattern $P$ in the text because in the other case we would have make a mistake changing the symbol $P[k]$ in the text. Therefore, we need to consider these two options in the transducer. In this way, there will be two transductions from the same state: the first one must read $P[k]$ and output the same symbol $P[k]$ and the second one  must read $P[k]$ but output $c.$  This leads us to the second and third parts of the transducer, which correspond to the two options that we have just described.

In our algorithm, lines from $2$ to $9$ basically describe the same algorithm presented in \cite{clrs} keeping track to the biggest found prefix so far. This for loop constructs the first and the second part of the transducer, where the input symbols are the same for the output for every edge. The state number $m = P.length$ will be always a \emph{sink state}, i.e. a state that has no out-going edge which means that the processed input is not going to be accepted by the transducer. The if sentence of the line $10$ is used to initialize the variable $transNode$ which will be the state where the edge with the transduction $P[k]/c$ is going to and where the third part of the transducer starts. The loop for in the line $14$ finishes the construction of the third part of the transducer. For each character in $P[k+1, \dots, m-2],$ it adds an edge with the same input and out symbol and an edge from each of these states to the sink state $m$ with the transduction $?/?.$ It means that if the pattern after the character $P[k]$ is not found, we have before made a mistake replacing $P[k]$ by $c$ and this current input cannot be accepted. Lines from $18$ to $20$ actually add the transition $P[k]/c$ and complete the third part of the transducer adding the edges from the last state of this part. Finally, the for loop in the line $21$ considers the cases when the transduction $P[k]/c$ must be realized from the states of the second part of the transducer verifying that this does not cause ambiguities. 

\begin{codebox}
\Procname{$\proc{Local-Extension}(P, k, c)$}
\li $m \gets \attrib{P}{length}$
\li \For $q \gets 0$ \To $m-1$
\Do
\li \For each character $a \in \Sigma$
\Do
\li $j \gets \min (m, q+1)$
\li \Repeat
\li \Do $j \gets j - 1$
\End
\li \Until $P_j \sqsupset P_{q-1} a$
\li $E \gets E \cup \{ (q, a, a, j+1) \}$
\End
\li $F \gets F \cup \{ q\}$
\End
\li \If $k = m-1$ \Do
	\li $transNode \gets 0$	
	\End
\li \Else \Do
	\li $transNode \gets m + 1$
	\li \For $i \gets k+1$ \To $m-2$ \Do
		\li $node \gets m + i - k$
		\li $E \gets E \cup \{ (node, P[i], P[i], node + 1 ) \}$
		\li $E \gets E \cup \{ (node, ?, ?, m ) \}$
		\End
	\End
\li $E \gets E \cup \{(k, P[k], c, transNode)\}$
\li $E \gets E \cup \{ (2m - k - 1, P[m-1], P[m-1], 0 ) \}$
\li $E \gets E \cup \{ (2m - k - 1, ?, ?, m ) \}$

\li \For $i \gets k+1$ \To $m-1$ \Do
	\li $S \gets P[k \twodots i -1] P[k \twodots m - 1]$
	\li \If $P_{k-1} \sqsupset P_{i-1} \text{ and } S_{m - k -1} \not\sqsubset _{i-k}S$ \Do
		\li $E \gets E \cup \{(i, P[k], c, transNode)\}$	
	\End
\End
\end{codebox}

\begin{figure}[!t]
\centering
\includegraphics[width=2.5in]{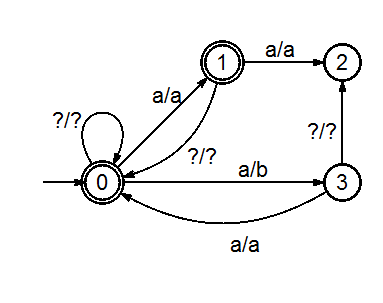}
\caption{Local extension for the rule $aa \rightarrow ba$.}
\label{fig_auto2}
\end{figure}

\begin{figure}[!t]
\centering
\includegraphics[width=2.9in]{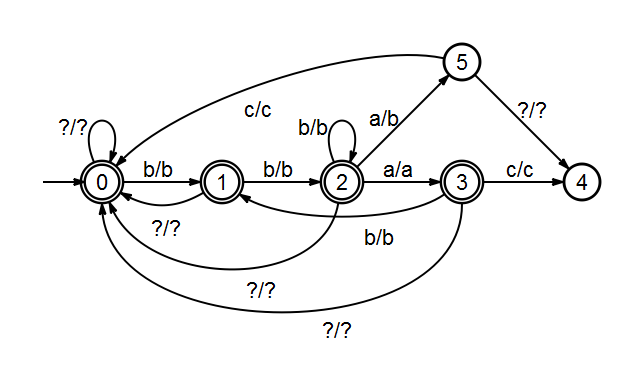}
\caption{Local extension for the rule $bbac \rightarrow bbbc$.}
\label{fig_auto3}
\end{figure}

In figure \ref{fig_auto2} the result of the algorithm for the same contextual rule of the transducer in  the figure \ref{fig_auto0} is shown. In figure \ref{fig_auto3} it is presented the local extension for the rule $bbac \rightarrow bbbc.$ The first part is composed of the states $0,1$ and $2.$ The second one, of the states $2, 3$ and $4.$ And the third one, of the sates  $2, 5 $ and $0.$ In this example, $m = 4$ is the sink state and the state $5$ is the $transState.$

\section{Experiments and Evaluation}
For training and testing, it was used the data set for Spanish provided in the CoNLL-$2002.$ The training file was divided in two parts: $85,1\%$ for the lexical tagger and $15,9\%$ for the contextual tagger. Since some trigger words could belong to some named entities, we decided to infer $100$ rules in the regular way, i.e. exploring among all possible rules, and then infer about $15$ rules which tried to replace those preposition tags by BIO tags. These $115$ rules produced a deterministic transducer of $15.828$ states.

The context length used in these rules was of size $\pm 2.$ We had made a previous experiment with size $\pm 3,$ but we noticed that it requires a huge amount of training time and the transducer size was also big because these longer rules were more complex and generated a more complex transducer when their local extension transducers were composed. However, the results with this configuration was similar to the $\pm2$ configuration, therefore we adopted to work with the last one configuration.

\begin{table}[!t]
\renewcommand{\arraystretch}{1.3}
\caption{Results for the data training sets.}
\label{table_example}
\centering
\begin{tabular}{c|c|c|c}
\hline 
data set & precision & recall & $F_{\beta = 1}$ \\ 
\hline 
Spanish dev. & $58,9\%$ & $54,5 \%$ & $56,6\%$ \\
\hline 
Spanish test  & $61,8\%$ & $58,3 \%$ & $60,0\%$ \\ 
\hline 
\end{tabular} 
\label{tab1}
\end{table}

Table \ref{tab1} presents the results of the system for the both data sets provided int the CoNLL-$2002.$ The final comparison used the \emph{test data set}. The \emph{development data set} was given to the participants for testing their models before the conference took place. Table \ref{tab2} describes the contextual tagger performance. The first line presents the measures for the case when we only used the lexical tagger. We can see that the support provided for the contextual tagger is around $10\%.$

\begin{table}[!t]
\renewcommand{\arraystretch}{1.3}
\caption{Performance of the contextual tagger.}
\label{table_example}
\centering
\begin{tabular}{c|c|c|c}
\hline 
 & precision & recall & $F_{\beta = 1}$ \\ 
\hline 
Before apply cont. rules & $50,3\%$ & $51,4 \%$ & $50,8\%$ \\
\hline 
After apply cont. rules  & $61,8\%$ & $58,3 \%$ & $60,0\%$ \\ 
\hline 
\end{tabular} 
\label{tab2}
\end{table}

\section{Discussion}
It was shown a different approach for the NEE task where a simple model totally based in automatons (or transducers) is learned. The time complexity required to identify and classify named entities in  text is lineal. In fact, the text is examined only two times: in the initialization stage using the data structure trie and in the next stage, the application of the transducer of the contextual rules. The results obtained are comparable with other works presented in the CoNLL-$2002$ with significant more powerful techniques. For instance, in \cite{mcnamee2002conll} is introduced a system based on linear support vector machines (SVMs) which had a precision of $56,28 \%$  and a recall $66,51 \%$, while in \cite{black2002conll} it was used techniques like transformation-based learning and decision trees obtaining a precision of $60,53\%$  and a recall of  $67,29\%.$ On overall, both systems attained a $F_1$ measure of $60,97 \%$ and $63,73\%$ respectively with the same data set.

Furthermore, as we have remarked before, the only assumption made in the system about the Spanish was the list of trigger words. This allow us to easily apply the system a different languages without make substantial changes. Therefore, we can still improve the proposed system as well as expand it to more languages.

\bibliographystyle{IEEEtran}
\bibliography{IEEEabrv,./paper}

\end{document}